\RequirePackage{fix-cm}
\documentclass[smallcondensed]{svjour3}     % onecolumn (ditto)
\smartqed  % flush right qed marks, e.g. at end of proof
\usepackage{graphicx}
\usepackage{amssymb}
\usepackage{amsmath}

\newcommand{\Borel}{\mathfrak B}
\newcommand{\Prob}{\mathsf P}
\newcommand{\Expect}{\mathsf E}
\newcommand{\Loss}{\mathcal L}
\newcommand{\Sample}{V}
\newcommand{\sample}{v}
\newcommand{\LSample}{\breve V}
\newcommand{\ERisk}{\widetilde R}
\newcommand{\EAvg}{\widetilde F}
\newcommand{\LRisk}{\breve R}
\newcommand{\WQ}{\widetilde Q}
\newcommand{\PsiThr}{\Upsilon}
\newcommand{\Capty}{L}
\newcommand{\URisk}{\bar R}

\journalname{Machine Learning}
\begin{document}

\title{Exact and empirical estimation of misclassification probability\thanks{
The work is supported by RFBR, grants 10-01-00113-a, 11-07-00346-a.}
}
%\subtitle{Do you have a subtitle?\\ If so, write it here}

%\titlerunning{Short form of title}        % if too long for running head

\author{Victor Nedelko}

%\authorrunning{Short form of author list} % if too long for running head

\institute{V. Nedelko \at
           Institute of Mathematics SB RAS, 4 Acad. Koptyug avenue, 630090 Novosibirsk, Russia\\
           Tel.: +7-913-926-8527\\
           Fax: +7-383-333-2598\\
           \email{nedelko@math.nsc.ru}           %  \\
%             \emph{Present address:} of F. Author  %  if needed
}

\date{Received: date / Accepted: date}
% The correct dates will be entered by the editor

\maketitle

\begin{abstract}
We discuss the problem of risk estimation in the classification problem,
with specific focus on finding distributions that maximize the confidence intervals of risk estimation.
We derived simple analytic approximations for the maximum bias of empirical risk for histogram classifier.
We carry out a detailed study on using these analytic estimates for empirical estimation of risk.
% \PACS{PACS code1 \and PACS code2 \and more}
% \subclass{MSC code1 \and MSC code2 \and more}
\end{abstract}

\section{Introduction}
\label{intro}

The study of overfitting is one of the most important research directions in the area of machine learning.
This problem arises from common disadvantage of more complex decision rules relative to the simpler ones
when the sample size is not very large.
In order to choose the optimal complexity of the method one needs to be able to estimate the quality of solution
without using the test sample.

In classification problems, or pattern recognition problems,
the quality of solutions is characterized by the misclassification probability (or by more general notion of risk).
Thus, what one needs is the ability to estimate risk as precisely and reliably as possible,
for given training sample and classification method.

The estimations of risk are carried out either in point or interval form.
In the former category there are the methods of empirical risk, cross validation estimation, bootstrap and so on.
The quality of point estimation is naturally characterized by the mean quadratic deviation from the estimated quantity.
This characteristic allows to compare different estimations and select the best.
It does not however provide sufficient information on how reliable the numerical estimation of risk in given problem is.
The latter requires the construction of interval form estimations.
Among these, the best known is the Vapnik--Chervonenkis estimation~\cite{VC74}.

There are many studies devoted to the construction and refinement of risk estimation,
conducted in the context of various approaches.
These approaches are well known, and we will not enumerate them here as many good reviews already exist~\cite{Langford}.
But the main reason why we do not discuss these approaches is that this paper is focused on a problem 
being substantially different from the subject of mentioned works.

The subject of this paper is not refinement of risk estimation in general case, but finding the exact (tight) bound
for special case.
Saying about an exact bound we mean that one can find an example where real risk is equal to estimated one.
So the Vapnik and Chervonenkis estimates are (nearly) tight because such example exists, see chapter~4 in~\cite{Langford}.

This means that one may improve VC--estimates only making some additional assumptions.
First, let us list the assumptions of original VC--approach:

1) The distribution the sample was drawn from is unknown.

2) The only statistic we may use is empirical risk.

3) The only information we have about the classification method is the complexity measure (VC-dimension or equivalent).

Under these conditions VC--estimates are (nearly) tight.

Instead of 3 we shall consider particular classification method, namely histogram classifier.

In this paper, we develop an approach based on the explicit finding of distributions
for which the error bounds of the estimations are maximum.
This does not mean however that we will be primarily oriented toward "the worst case" scenario
in terms of expected quality of classification.
As is well known, analytically constructed "worst" cases are seldom realized in practice
(well known example is the simplex method that quickly solves most practical problems,
although there are examples that require exponential time).
However, large error bounds do not imply high risk, and the distributions with maximal error bounds are by no means "bad".
Rather to the contrary, it is with these "bad" distributions
that we can very accurately estimate the probability of misclassification.

Note that the Vapnik--Chervonenkis estimation is unduly pessimistic because of the fact
that they "focus on the worst case scenario".
The key point here is not that one assumes the "worst case" distribution,
but rather that any classification method, including those whose classifiers are maximally different
(not taking into account the "similar classifier" effect) are allowed~\cite{Vorontsov}.

As will be shown below, the distribution that delivers maximum bias of empirical risk for the histogram classifier
(the bias in the estimation is the main contributor to its error) is quite typical.

The histogram classifier is convenient for the study because it allows us to make analytical calculations~\cite{BD05},
provides simplicity in the method of classification,
and does not have any extra features which could potentially improve estimates.

In~\cite{Nedelko03} the maximum bias of the empirical risk of the histogram classifier was obtained in the asymptotic case.
The result is based on the proof of the assertion that the maximum bias of empirical risk is achieved
by a piecewise-constant distribution.
The assumption was made of distribution having no more than three areas of constancy,
which makes it possible to find such a distribution by numerical optimization.

In this paper, the result is extended to finite samples.
We also present the explicit form of the distribution, for which we obtain a maximum bias of the empirical risk.
Finally, we find simple formulas that approximate the needed dependency with sufficient precision.

We study the possibility of using these results to obtain empirical estimations of risk.
In this context, we define as empirical the estimations, the properties of which are not proven but established empirically,
using in particular, statistical modeling methods.
Note that the estimations that are most widely used in practice are empirical.
Take as an example the estimation by the cross validation:
its error bound has not been theoretically estimated in general case,
but practical experience of solving problems allows to recommend this estimation.

\section{Formulation of the problem}
\label{sec:1}
%Text with citations \cite{RefB} and \cite{RefJ}.
%\subsection{Subsection title}
%\label{sec:2}
%as required. Don't forget to give each section
%and subsection a unique label (see Sect.~\ref{sec:1}).

% For one-column wide figures use
%\begin{figure}
% Use the relevant command to insert your figure file.
% For example, with the graphicx package use
%  \includegraphics{example.eps}
% figure caption is below the figure
%\caption{Please write your figure caption here}
%\label{fig:1}       % Give a unique label
%\end{figure}
%
% For two-column wide figures use
%\begin{figure*}
% Use the relevant command to insert your figure file.
% For example, with the graphicx package use
%  \includegraphics[width=0.75\textwidth]{example.eps}
% figure caption is below the figure
%\caption{Please write your figure caption here}
%\label{fig:2}       % Give a unique label
%\end{figure*}

Let us consider the general formulation of the problem of decision rule construction
(pattern recognition, supervised classification).

Let $X$ be the space of possible values of features (predictors) 
and $Y$ -- the space of possible values of features to be predicted.
Let $C$ be the set of all probabilistic measures on a given $\sigma$-algebra of subsets of $D=X\times Y$.
For each $c\in C$ there is a probabilistic space $\langle D,\Borel,\Prob_c\rangle$
where $\Borel$ is a $\sigma$-algebra, and $\Prob_c$ a probabilistic measure.
Subscript $c$ was introduced for further usage as a short notation of certain probabilistic measure.

A decision function (rule) is defined as $\lambda\colon X\to Y$.

A quality of a decision function is measured by the loss function:  $\Loss\colon Y^2\to [0,\infty )$.

A risk is defined as the expected (average) loss
$$
R(c,\lambda )=\Expect\Loss (y,\lambda (x)) =\int\Loss (y,\lambda (x)) \Prob(dx,dy).
$$

In this paper we use the simplest loss function
$\Loss(y,y^\prime)= \left\{
\begin{smallmatrix}0,& y=y^\prime\\ 1,& y\neq y^\prime
\end{smallmatrix}\right.$.
In this case a risk is a misclassification probability.

Note that risk depends on a distribution that is unknown, so it is necessary to estimate a risk with a sample.

Let $\Sample_N =\left((x^i,y^i)\in D\,\vert \,i=1,\dots,N\right)$
be a random independent identically distributed (i.i.d.) sample from distribution $\Prob_c$,  $\Sample_N\in D^N$.
In most cases the size of the sample $N$ will be fixed, so we will drop this subscript in the notation of the sample.

There are many single-value risk estimates: empirical risk (resubstitution error), leave-one-out estimate (special case of cross-validation), bootstrap etc.

We define empirical risk as the average loss on the sample:
$$
    \ERisk(\Sample, \lambda)={1\over N}\sum_{i=1}^N \Loss(y^i,\lambda(x^i)).
$$

Let $Q\colon D^N\to\Lambda$ be an algorithm (method) for constructing decision functions
and $\lambda_{Q,\Sample}$ -- the function constructed on sample $V$ by $Q$.
Here $\Lambda$ is a given class of decision functions.

The Leave-one-out estimate is defined as
$$
\LRisk(\Sample, Q)={1\over N}\sum_{i=1}^N \Loss(y^i,\lambda_{Q,\LSample^i}(x^i)),
$$
where $\LSample^i = \Sample \backslash \{(x^i,y^i)\}$ is a sample
produced from $\Sample$ by removing the $i$-th observation.

This estimate is nearly unbiased since
$$
\Expect \LRisk(\Sample_N, Q) = \Expect R(c,\lambda_{Q,\Sample_{N-1}}),
$$
where an expectation is taken over all samples of given size ($N$ and $N-1$ correspondently).

Although leave-one-out rate is a nearly unbiased risk estimate,
we may not use it immediately as an expected (predicted) risk value.
For example, if we get $\LRisk=0$,
then we have no reason to expect that the misclassification probability on new objects will be zero.
By estimating misclassification probability as null we assert that the classifier will not make errors on new objects.
However, in the general case, it is not possible to prove this statement by a finite sample.

In general, a single-value risk estimate is some function of the sample.

A method $\WQ$ that provides
$\lambda_{\WQ,\Sample} = \arg\min_{\lambda\in\Lambda} \ERisk(\Sample, \lambda)$
is called an empirical risk minimization method.

A bias of an empirical risk is defined as
$$
S(c) = F(c)-\EAvg(c),
$$
where $F(c)=\Expect R(c,\lambda_{Q,\Sample})$,
$\EAvg(c)=\Expect \ERisk(\Sample,\lambda_{Q,\Sample})$.
Note that both the means in $F(c)$ and $\EAvg(c)$ are made on all possible decision functions,
since the dependency of $\EAvg$ on $c$ is implicitly included in $V$.

The maximal bias of empirical risk is
\begin{equation}
\label{eq:bias}
\hat S (\EAvg_0) = \max_{c\colon\EAvg(c)=\EAvg_0} S(c).
\end{equation}
Here $\EAvg_0$ is some chosen value of empirical risk.

Note that we are interested in a conditional maximum,
because in practice we have a certain value of empirical risk calculated for given data.
Then by substituting this value as $\EAvg_0$ we get a reasonable estimate of maximal risk bias for the data.

%An equivalent task is finding
%$$
%\hat S^\prime (F_0) = \max_{c\colon F(c)=F_0} S(c).
%$$
%
%It is obvious that $\hat S (\EAvg_0) = \hat F (\EAvg_0) - \EAvg_0$,
%where $\hat F (\EAvg_0) = \max_{c\colon\EAvg(c)=\EAvg_0} F(c)$,
%and also $\hat F^{-1} (F_0) = \check F (F_0)$,
%where $\check F (F_0) = \min_{c\colon F(c)=F_0} \EAvg(c)$.
%Here $\hat F^{-1} (\cdot)$ is the inverse function.
%The last form is useful, because $\check F (F_0)$ allows very simple approximation (see below).

\section{Histogram classifier}

\subsection{Description}

Let $X$ be discrete, i.e. $X = \{1, \dots, k\}$ and decision function minimizes an empirical risk in each $x\in X$.
In this case a probabilistic measure $c\in C$ can be defined by a matrix of parameters 
$$
c=\left(c_j \vert j=1,\dots,k \right), \;\;\;\; c_j=(\alpha_j,p_j),
$$
where $\alpha_j=\Prob(x=j)$, $p_j=\Prob(y=1\vert x=j)$.

%In this case a probabilistic measure $c\in C$ can be defined by a set of probabilities 
%$$
%c=\{\zeta^\omega_j = \Prob(x=j,y=\omega)\vert j=1,\dots,k,\omega=0,1\}.
%$$
%Let us denote $\alpha_j=\Prob(x=j)=\zeta^0_j+\zeta^1_j$,
%$p_j=\Prob(y=1\vert x=j)$,
%$q_j=1-p_j$,
%$c_j=(\alpha_j,p_j)$.

For a sample $\Sample$ of size $N$ let $n_j$ be a number of sample points with $x=j$ and
$m_j$ be a number of points with $x=j$ and $y=1$.
So a sample is defined as a set of pairs $v_j=(m_j,n_j)$, i.e. $\Sample=\{v_j\vert j=1,\dots, k\}$.
To describe a sample we shall also sometimes say that in a "cell" $j$ there are $m_j$ points of the first class
and $n_j-m_j$ points of the zero class.

Consider the algorithm $\WQ$ that minimizes an empirical risk in each $x\in X$,
i.e. $\lambda_{\WQ,\Sample}(j)=0$, when $n_j-m_j>m_j$, $\lambda_{\WQ,\Sample}(j)=1$, when $n_j-m_j<m_j$,
and $\lambda_{\WQ,\Sample}(j)$ takes randomly 1 or 0, when $n_j-m_j=m_j$.

There is polynomial distribution on samples that allows analytical averaging of some sample functions.

\subsection{Expectation of additive functions}

In order to find the risk bias we need to calculate risk and empirical risk expectations.

Let $f(\Sample,c) = \sum_{j=1}^k \varphi(v_j,c_j) = \sum_{j=1}^k \varphi(m_j,n_j,\alpha_j,p_j)$
be an additive function of a sample and of a distribution.
Then $\Expect f(\Sample,c) = \sum_{j=1}^k \Expect\varphi(v_j,c_j)$.

Denote $B(m,n,p) = C_n^m p^m (1-p)^{n-m}$ -- Binomial distribution.

Let us introduce $\mu_\varphi(c) \equiv \mu_\varphi(\alpha,p) = \Expect\varphi(\sample,c)$.
%It is easy to see that 

We can directly (or by summation of multinomial distribution) obtain
\begin{multline}
\label{eq:cellavg}
\mu_\varphi(\alpha,p) = \sum_{n=0}^N B(n,N,\alpha) \sum_{m=0}^n B(m,n,p) \varphi(m,n,\alpha,p) = \\
= \sum_{n=0}^N B(n,N,\alpha) \pi_\varphi(n,\alpha,p),
\end{multline}
where $\pi_\varphi(n,\alpha,p) = \sum_{m=0}^n B(m,n,p) \varphi(m,n,\alpha,p)$.

Finally,
$$
\Expect f(\Sample,c) =  \sum_{j=1}^k \mu_\varphi(c_j).
$$

An empirical risk and misclassification probability are additive, namely:
\begin{align*}
\ERisk(\Sample) & = \sum_{j=1}^k \tilde r(m_j,n_j),  \\
\tilde r(m,n) = {1\over N}\tilde \nu(m,n), & \;\;\;\;    \tilde \nu(m,n) =\min(m,n-m); \\
R(c,\lambda_{\WQ,\Sample}) & = \sum_{j=1}^k r(m_j,n_j,\alpha_j,p_j), \\
r(m,n,\alpha,p) = \alpha \nu(m,n,p), & \;\;\;\;
\nu (m,n,p) = \left\{\begin{smallmatrix} 1-p, & m>n-m \\ p, & m<n-m \\ 0.5, & m=n-m \end{smallmatrix}\right..
\end{align*}
   
\subsection{Finding maximal bias}

The task~\eqref{eq:bias} for histogram classifier case has the form:
\begin{equation}
\label{eq:discrete}
\sum_{j=1}^k \mu_s(\alpha_j,p_j) \to \max_{\alpha_j,p_j},
\end{equation}
with constraints:
$$
\sum_{j=1}^k \mu_{\tilde r}(\alpha_j,p_j) = \EAvg_0, \;\;\;\; \sum_{j=1}^k \alpha_j = 1,  \;\;\;\; 0\leq p_j\leq 1.
$$

Here $\mu_s(\alpha,p)$ and $\mu_{\tilde r}(\alpha,p)$ are defined by formula~\eqref{eq:cellavg}
with correspondent substitution $s(m,n,\alpha,p) = r(m,n,\alpha,p) - \tilde r (m,n)$ and $\tilde r (m,n)$
instead of $\varphi (m,n,\alpha,p)$.

\begin{figure}[t]
    \centering
    \includegraphics[width=100mm]{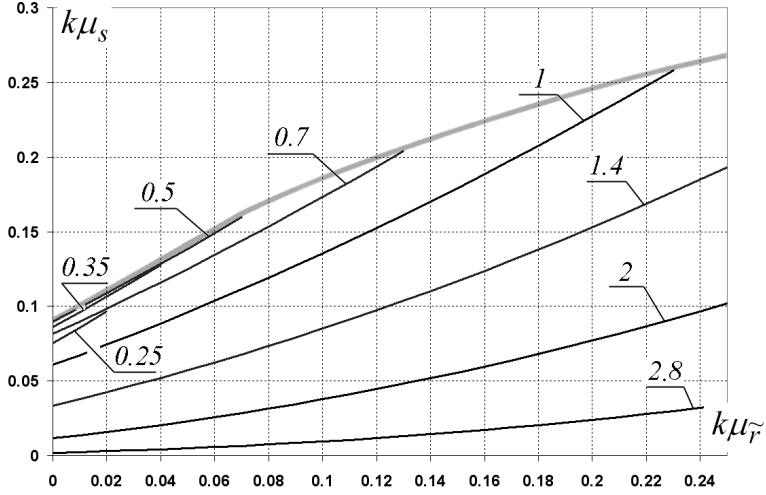}
    \caption{Dependencies for risk on a "cell".}
    \label{figRbyRe}
\end{figure}

Figure\,\ref{figRbyRe} shows dependences $k\mu_s$ on $k\mu_{\tilde r}$ for different $k\alpha$, when $N=20$, $k=10$.

%where $\hat\mu_s(z) = \max_{\mu_{\tilde r}=z}\mu_s$ is the envelope of the plotted curves.
Let $\zeta(z) = k \max_{k \mu_{\tilde r}=z}\mu_s$ be the envelope of the plotted curves.
On figure\,\ref{figRbyRe} it is shown as thick grey curve.

One can show that the maximal bias differs from $\zeta(\EAvg_0)$ less than by ${1\over k} + 0{.}01\cdot \zeta(\EAvg_0)$.
The proof is based on the following.

Let us consider the task~\eqref{eq:discrete} without the constraint $\sum_{j=1}^k \alpha_j = 1$.
Then it may be transformated to the equivalent form:
\begin{equation}
\label{eq:noconstraint}
\sum_{j=1}^k \zeta(z_j) \to \max_{z_j},
\end{equation}
with constraints: $\sum_{j=1}^k z_j = \EAvg_0$, $0\leq z_j\leq 1$.

%Note that ${\hat\mu_s(z)\over z}$ is monotonously nonincreasing when $z$ is increasing.
Suppose $\zeta(z)$ to be concave, i. e. $\zeta^{\prime\prime}(z)<0$.
Then the maximum in~\eqref{eq:noconstraint} is attained when the all $z_j$ are equal.
Hence $\zeta(\EAvg_0)$ would be a solution of problem~\eqref{eq:discrete} without the constraint $\sum_{j=1}^k \alpha_j = 1$.
The maximal bias appears when the all $\alpha_j$ are equal, but their sum may be less than $1$.

However one can easily keep the constraint $\sum_{j=1}^k \alpha_j = 1$ by setting all extra probability to any "cell",
and setting correspondent $p_j=0$ so as to provide $\EAvg$ to be unchanged.

In reality $\zeta(z)$ is not concave, but may be very tightly approximated by a concave function.
Numerical evaluation shows that the relative error caused by this approximation is less than~$0{.}01$.

The form of the envelope curve on figure\,\ref{figRbyRe} is typical for $N\geq k$.
The left part of the envelope corresponds to $p$ changing from $0$ to $0.5$ when $\alpha={1\over N}$.
The right part of the envelope corresponds to $\alpha$ changing from ${1\over N}$ to ${1\over k}$ when ${p=0{.}5}$.

The case $N<k$ is simpler and isn't so important in practice, so we do not consider it here.
For simplicity let's assume $N\geq k$.

Thus the distribution providing a bias that differs from the supreme bias less than by ${1\over k}$ is of the following form:
\begin{gather*}
\alpha_j=\alpha^\prime, \; p_j=p^\prime, \;\; j=1,\dots,k-1, \\
\alpha_k=1-{k-1\over N}, \; p_k=0.
\end{gather*}

In other words, the "worst" distribution is uniform over all the "cells"
except for one cell that accumulates the "excess" probability.

The values $\alpha^\prime$ and $p^\prime$ are determined by the following way.

When $\EAvg_0\leq\EAvg_T$ we have $\alpha^\prime={1\over N}$, and $p^\prime$ is calculated via $\EAvg_0$.
When $\EAvg_0\geq\EAvg_T$ we have ${p^\prime=0{.}5}$, and $\alpha^\prime$ is calculated via $\EAvg_0$.
Here $\EAvg_T$ is the expectation of empirical risk by  ${\alpha^\prime={1\over N}}$,\: $p^\prime=0{.}5$.

\subsection{Approximations}

Binomial distribution in~\eqref{eq:cellavg} may be tightly approximated by Poisson distribution
$$
\mu_\varphi(\alpha,p) \approx \rho_\varphi(\gamma,p) = \sum_{n=0}^N {\gamma^n\over n!}e^{-\gamma} \pi_\varphi(n,\alpha,p),
$$
where $\gamma=N\alpha$.

Since $\pi_{\tilde r}(n,\alpha,p)={1\over N}\pi_{\tilde \nu}(n,p)$
and ${\pi_{r}(n,\alpha,p)=\alpha\pi_{\nu}(n,p)}$ we have
\begin{gather*}
\rho_{\tilde r}(\gamma,p) = {1\over N} \rho_{\tilde \nu}(\gamma,p);\\
\rho_{r}(\gamma,p) = \alpha \rho_{\nu}(\gamma,p) = {\gamma\over N}\rho_{\nu}(\gamma,p).
\end{gather*}
Functions $\nu$ and $\tilde \nu$ are defined in section 3.2.

Since for the "worst" distribution $c^*$ the probabilities in all "cells" are the equal
except for the last "cell" that does not affect the risks, we obtain
\begin{gather*}
\EAvg(c^*)\approx {k-1\over N} \rho_{\tilde \nu}(\gamma,p) \approx {1\over M} \rho_{\tilde \nu}(\gamma,p), \\
F(c^*)\approx {(k-1)\gamma\over N}\rho_{\nu}(\gamma,p) \approx {\gamma\over M}\rho_{\nu}(\gamma,p),
\end{gather*}
where $M={N\over k}$ is the average number of sample points per cell, i. e. is the relative sample size.
 
Let $\tilde\rho_1 (p) = \rho_{\tilde \nu} (1,p)$ and $\tilde\rho_2 (\gamma) = \rho_{\tilde \nu} (\gamma,\tfrac12)$.

Let us introduce the function
$$
\psi(z)=
\begin{cases}
\psi_1(z), &0\leq z\leq \PsiThr;\\
\psi_2(z), &z\geq \PsiThr,
\end{cases}
$$
where
$\PsiThr=\rho_{\tilde \nu}(1,\tfrac12)\approx 0{,}163$,\: 
$\psi_1(z)=\rho_{\nu}\bigl(1,p^*(z)\bigr)$,\: 
$p^*(z)=\tilde\rho_1^{-1}(z)$,\: 
$\psi_2(z)=\gamma^*(z)\rho_{\nu}\bigl(\gamma^*(z),\tfrac12\bigr)=\tfrac12 \gamma^*(z)$,\: 
$\gamma^*(z)=\tilde\rho_2^{-1}(z)$.

When $M\geq 1$ the approximate expression for maximal bias of empirical risk is
$$
\hat S(\EAvg_0) \approx {1\over M}\psi(M\EAvg_0) - \EAvg_0.
$$
%where $\psi'(z)=\psi(z)-z$.
This function is defined for those $\EAvg_0$ that provide for the condition ${1\over M}\psi(M\EAvg_0)\leq 0{.}5$.

\begin{figure}[t]
    \centering
    \includegraphics[width=110mm]{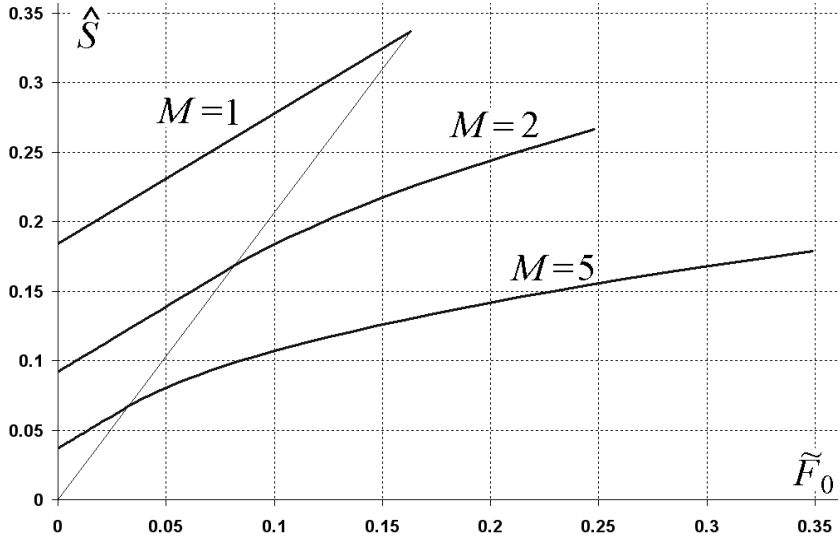}
    \caption{The maximal bias of empirical risk.}
    \label{figBias}
\end{figure}

Figure\,\ref{figBias} shows the maximal bias of empirical risk for a different value $M={N\over k}$.
The thin line contains the points of joining $\psi_1$ and $\psi_2$.

The function $\psi(z)$ can be easily calculated numerically,
however it may be approximated by simple analytical expressions.

As fig.\,\ref{figBias} shows, the function $\psi_1(z)$ is close to linear.
Since $\psi_1(0)={1\over 2e}$ and $\psi_1(\PsiThr)=0{.}5$ one gets a linear approximation
$\psi_1(z)\approx \bar\psi_1(z)={1\over 2e}+\left ({1\over 2}-{1\over 2e}\right ){z\over \PsiThr}$.

The function that is inverse to $\psi_2(z)$ can also be easily approximated.
It appears that
$1-2{\tilde\rho_2 (\gamma)\over\gamma} \approx {1-2\PsiThr\over\sqrt{1+2\gamma}}\sqrt{3}$.

Taking into account $\psi_2(z)=0{,}5\tilde\rho_2^{-1}(z)$ or $\psi_2^{-1}({\gamma\over 2})=\tilde\rho_2(\gamma)$
and putting ${t={\gamma\over 2}}$ one obtains
$\psi_2^{-1}(t) \approx \bar\psi_2^{-1}(t) = t\cdot\left(1- {1-2\PsiThr\over\sqrt{1+4t}}\sqrt{3}\right)$,\:
$t\geq{1\over 2}$.

Finally, by combining $\bar\psi_1(z)$ and $\bar\psi_2(z)$ we obtain
the approximation $\bar\psi(z)$ for $\psi(z)$.
The relative error $\Delta(z)={\bar\psi(z)-\psi(z)\over\psi(z)}$ of this approximation does not exceed $0.01$.

\section{Estimates comparison}

Now we are to compare the obtained exact estimates of empirical risk bias
with the complexity based estimates by Vapnik and Chervonenkis.

Let us denote
$H(\tilde p,p) = \tilde p\ln{\tilde p\over p} + (1-\tilde p)\ln{1-\tilde p\over 1-p}$ -- an entropy,
$\Capty$ -- a number of decision rules in the set $\Lambda$.

When $N\to\infty$, $\varkappa={N\over\ln \Capty}=const$ the VC-estimates are determined from equation
\begin{equation}
\label{eq:vc_est}
H(\EAvg_0,\hat F_{VC}(\EAvg_0)) = {1\over\varkappa}
\end{equation}

\begin{figure}[t]
    \centering
    \includegraphics[width=100mm]{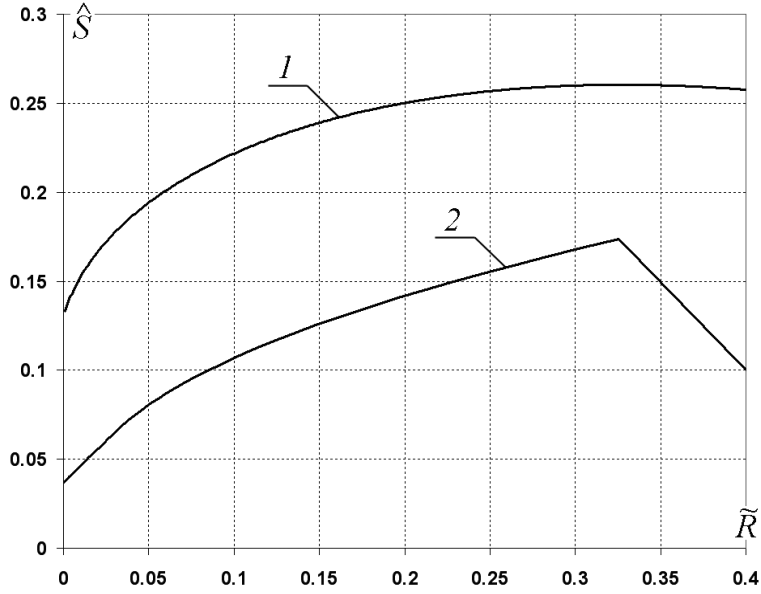}
    \caption{Comparison VC-estimate with exact bias estimate.}
    \label{figComparison}
\end{figure}

The solution $\hat F_{VC}(\EAvg_0)$ of this equation appears to be unimproved (in general case)
asymptotical estimate of risk based on empirical risk.

For histogram classifier we have $\Capty=2^k$ or $\varkappa={N\over k\ln 2}$.
By substitution it into~\eqref{eq:vc_est} one obtain risk estimate that may be written in the form of bias
$$
\hat S_{VC}(\EAvg_0) = \hat F_{VC}(\EAvg_0) - \EAvg_0.
$$

Figure\,\ref{figComparison} shows the plots: 1 -- VC-estimate $\hat S_{VC}(\EAvg_0)$
and  2 -- exact bias estimate $\hat S(\EAvg_0)$ by ${N\over k}=5$.
This comparison provides apprehension on tightness of VC-estimates.

\section{Numerical evaluation}

In the case of continuous features space it is possible to construct probability distributions
that are "similar" to the distributions that provide maximal bias of empirical risk for histogram classifier.

We set out to analyze the distributions that maximize the empirical risk bias.
The distributions are obtained by maximizing the expected risk by fixed $\EAvg_0$
or by minimizing the expected empirical risk by fixed expected risk.

Consider first the change in distribution that delivers the maximal bias in histogram classifier case
in respect to changes in $F_0$.

When $F_0=0.5$ the minimum of $\EAvg$ is reached for the uniform distribution on $D$,
i.e. by $\alpha_j = {1\over k}$, $p_j = 0.5$.

When $F_0$ decreases all the probabilities $p_j$ stay equal to $0.5$,
except one of them, say $p_k$, that has to be $0$ or $1$.
The probability $\alpha_k$ that is redistributed to this "cell" increases according to the reduction of $F_0$.

When $F_0$ decreases further the $\alpha_k$ grows and $\alpha^\prime$ decreases to $1\over N$.
Then $\alpha^\prime$ stops decreasing, but $p^\prime$ starts changing.

Note that resulting distributions are characterized by the fact
that the space $X$ is split into two subsets: the first has zero Bayesian risk level,
and the second for which the Bayesian misclassification probability is substantial.

Such peculiar feature of the distribution can be easily provided in continuous space as well.

Let $X=[0,1]^n$ be $n$-dimensional hypercube with uniform probabilistic distribution.

To specify probabilistic measure $\Prob$ on $D$ one also needs to assign 
$g(x)=\Prob(y=1\vert x)$ -- conditional probability that an object belongs to the first class when its coordinates are $x$.

Let us construct $g(x)$ in the form
$$
g(x)=
\begin{cases}
g_1, & x_j<\delta, j=1,\dots ,n\\
g_2, & otherwise
\end{cases}, \;\;\;\;\;\;
\delta=\vartheta^{1\over n}.
$$

In another words, $g(x)$ is piece-wise constant with the two areas of constancy:
hypercube of volume $\vartheta$, and it's supplement to the hypercube of volume $1$.

Let us assign the first subclass of distributions (call them as model A) by the following:
$g_2=1$; parameter $\vartheta$ is equal to some constant $\vartheta_0$
and $g_1$ varies from $0$ to $0.5$; or $g_1=0.5$ and $\vartheta$ varies from $\vartheta_0$ to $0$.

Note that this distributions are constructed "by similarity pattern" of distributions
those provide maximal bias of empirical risk for histogram classifier.

\begin{figure}[t]
    \centering
    \includegraphics[width=110mm]{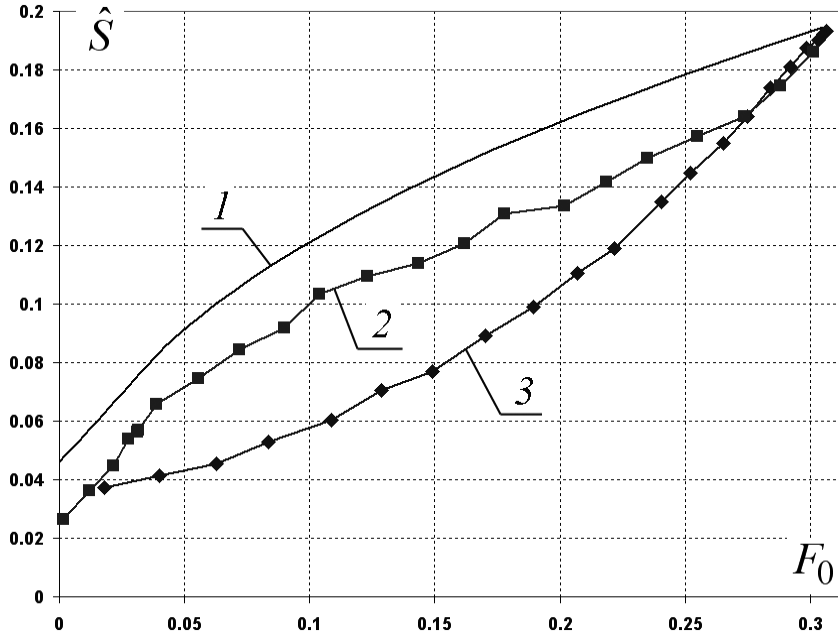}
    \caption{Empirical risk bias estimated via statistical simulation.}
    \label{figEmp_bias}
\end{figure}
 
For the sake of comparison we shall also consider one more subclass of distributions (call it as model B),
that is defined by the following: $\vartheta=0.5$, $g_1=g^\prime$, $g_2=1-g^\prime$,
where $g^\prime$ is a Bayesian risk level and varies from $0$ to $0.5$.

Figure\,\ref{figEmp_bias} shows the results of statistical simulation of classification by decision trees.
The greedy method of tree construction was used~\cite{Quinlan}.
Here: 1 -- the estimate $\bar\psi(z)$ for $M=4$;
2 -- modeling on distributions of model A with $\vartheta_0=0.83$,
3 -- modeling on distributions of model B.

The distributions from the model A delivers greater risk bias than any other distribution that have been examined.

\section{Empirical confidence intervals}

The constructed classes of distributions can be useful for construction of empirical confidence intervals~\cite{Nedelko09}
for risk.

A confidence interval for $R$ will be in the form $[0,\hat R(\Sample)]$.

We use only one-sided estimates because usually one has no need in the lower-bound risk estimates.
So the construction of a confidence interval is equivalent to choosing function $\hat R(\Sample)$
that will be called the estimating function.

For $\hat R(\Sample)$ the following condition must be held for any $c$:
\begin{equation}
\label{eq:emp_conf}
\Prob(R\leq\hat R(\Sample))\geq\eta
\end{equation}
where
$\eta$ is a given confidence probability.

Note that a confidence interval is built for given algorithm $Q$.

Known risk estimates are usually constructed not as immediate functions of a sample,
but via superposition $\hat R(\Sample) = R_e(\URisk(\Sample))$,
i.e. as a function of some empirical functional $\URisk(\Sample)$,
that may be, for example, empirical risk or leave-one-out rate.

Empirical functional here plays a role of a point estimate, that being a base to construct an interval estimate.

Analytical estimation of a confidence probability appears to be problematical in practice,
since it requires calculating the infimum over the all distributions (all probabilistic measures on $\Borel$),
therefore constructing empirical bounds is quite desirable.

By empirical estimation here we mean the estimating function that is obtained
via the estimation of the minimal confidence probability over some heuristically chosen finite set of distributions.
If such set is "wide" enough, one may believe that the estimation is valid.
As one has been unable to find a distribution that violates the estimate through a dedicated effort,
one can expect that the distribution in real world examples will not violate the confidence bound either.

\begin{figure}[t]
    \centering
    \includegraphics[width=80mm]{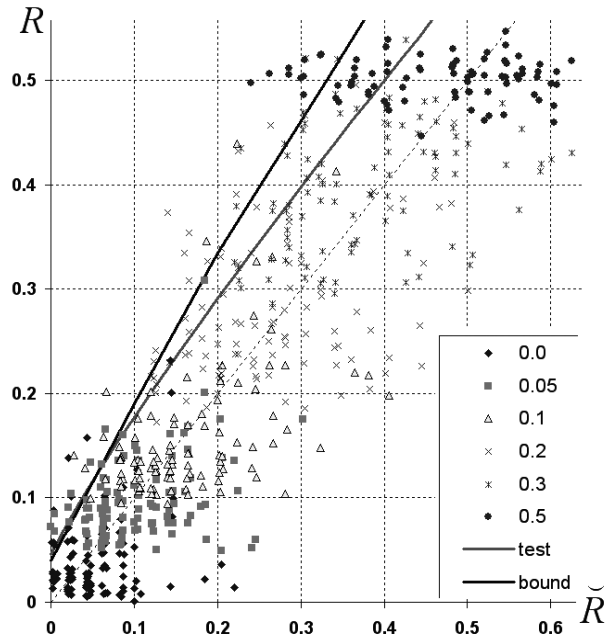}
    \caption{Empirical bounds for risk by $\eta=0.9$, built by leave-one-out estimate.}
    \label{figEmp_int}
\end{figure}

On figure\,\ref{figEmp_int} the empirical bounds for risk are shown when the number of terminal nodes in tree is equal to $3$,
sample size $N = 50$ and space dimensionality $n = 2$.
As distributions for modeling the parametrical set of distributions was chosen,
where $\Prob$ is uniform in the hypercube $[0,1]^n$
and  $\Prob(y=1\vert x)=
\begin{cases}
g^\prime, & x\in [0,\delta]^n\\
1-g^\prime, & x\not\in [0,\delta]^n
\end{cases}$.
Here $\delta=2^{1\over n}$ and the parameter $g^\prime$ defines a Bayesian error value
which is equal in this case to misclassification probability for an optimal decision tree.

\section{Conclusions}

In this paper, we considered the problem of constructing estimations for the probability of misclassification.
We found the maximum bias of empirical risk for the histogram classifier.
We found simple approximation formulas for this quantity.

We established that for histogram classifier,
the "worst" distribution (which maximizes the bias of empirical risk) is a mixture of the uniform (in $X$)
and an impulse distribution (concentrated in one point).

Similar type of distributions can be easily constructed in the space of continuous variables.
Results of statistical modeling for the problem of classification with decision trees suggest
that the maximum bias of empirical risk can be attained as well with other (besides the histogram classifier)
methods of classification, allowing, in particular,
to build more accurate empirical confidence intervals for the risk.

%\begin{acknowledgements}
%If you'd like to thank anyone, place your comments here
%and remove the percent signs.
%\end{acknowledgements}

% Non-BibTeX users please use


\begin{thebibliography}{}
%
% and use \bibitem to create references. Consult the Instructions
% for authors for reference list style.
%
%\bibitem{RefJ}
% Format for Journal Reference
%Author, Article title, Journal, Volume, page numbers (year)
% Format for books
%\bibitem{RefB}
%Author, Book title, page numbers. Publisher, place (year)

%\bibitem[Vapnik and Chervonenkis(1974)]{VC74}
\bibitem{VC74}
Vapnik V. N., Chervonenkis A. Ya. Theory of pattern recognition. M.: "Nauka". 1974. 415 p. (in Russian).

%\bibitem[Quinlan(1986)]{Quinlan}
\bibitem{Quinlan}
Quinlan J. Induction of decision trees, Machine Learning. 1986. Vol. 1, No. 1. Pp. 81-106.

%\bibitem[Braga-Neto and Dougherty(2005)]{BD05}
\bibitem{BD05}
Braga-Neto U. and Dougherty E.R. Exact performance of error estimators for discrete classifiers
// Pattern Recognition, Elsevier Ltd. - 2005. - V. 38, N. 11. - Pp. 1799-1814.

%\bibitem[Langford(2002)]{Langford}
\bibitem{Langford}
Langford J. Quantitatively tight sample complexity bounds. Carnegie Mellon Thesis. - 2002.
 - http://citeseer.ist.psu.edu/langford02quantitatively.html. - 130 p.

%\bibitem[Nedelko(2003)]{Nedelko03}
\bibitem{Nedelko03}
Nedelko V.M. Estimating a Quality of Decision Function by Empirical Risk
// LNAI 2734. Machine Learning and Data Mining in Pattern Recognition.
Third International Conference, MLDM 2003, Leipzig. Proceedings. Springer-Verlag. pp. 182-187.

%\bibitem[Nedelko(2009)]{Nedelko09}
\bibitem{Nedelko09}
Nedelko V.M. Empirical bounds for risk in some machine learning tasks.
// The XIII-th Int. Conference "Applied Stochastic Models and Data Analysis".
ASMDA–2009. Vilnius, Lithuania. 2009. P.115--119.

%\bibitem[Vorontsov(2008)]{Vorontsov}
\bibitem{Vorontsov}
Vorontsov K.V. Combinatorial probability and the tightness of generalization bounds
// Pattern Recognition and Image Analysis. - 2008. - V. 18, N. 2. - Pp. 243-259.


\end{thebibliography}
\end{document}